\documentclass[letterpaper, 10 pt, conference]{ieeeconf}
\IEEEoverridecommandlockouts
\usepackage{cite}
\usepackage{amsmath,amssymb,amsfonts}
\usepackage{algorithmic}
\usepackage{graphicx}
\usepackage{textcomp}
\usepackage{xcolor}
\usepackage{url}
\usepackage{siunitx}
\usepackage{mathrsfs}
\usepackage{bm}
\usepackage{mathtools}
\usepackage{siunitx}
\def\BibTeX{{\rm B\kern-.05em{\sc i\kern-.025em b}\kern-.08em
    T\kern-.1667em\lower.7ex\hbox{E}\kern-.125emX}}

\usepackage[acronym,shortcuts]{glossaries}
\newacronym{HILO}{HILO}{human-in-the-loop optimization}
\newacronym{MSE}{MSE}{mean squared error}
\newacronym{RP}{RP}{retinitis pigmentosa}
\newacronym{DSE}{DSE}{deep stimulus encoder}
\newacronym{SPV}{SPV}{simulated prosthetic vision}
\glsdisablehyper

\begin{document}

\title{\LARGE \bf Evaluating Deep Human-in-the-Loop Optimization for Retinal Implants Using Sighted Participants

\author{Eirini Schoinas$^{1}$, Adyah Rastogi$^{2}$, Anissa Carter$^{3}$, Jacob Granley$^{2}$, and Michael Beyeler$^{2,4}$}
\thanks{$^{1}$ES is with the College of Creative Studies at the University of California, Santa Barbara, CA 93106, USA.}
\thanks{$^{2}$AR, JG, and MB are with the Department of Computer Science, University of California, Santa Barbara, CA 93106, USA.}
\thanks{$^{3}$AC is with the Department of Communication, University of California, Santa Barbara, CA 93106, USA.}
\thanks{$^{4}$MB is with the Department of Psychological \& Brain Sciences, University of California, Santa Barbara, CA 93106, USA.}
\thanks{The authors confirm contribution to the paper as follows. 
Study conception and design: ES, AC, JG, MB; 
computational model development: JG, MB; 
experiment coding and development: ES, AR, AC, JG;
data collection: AC, ES, AR;
all authors analyzed the data, wrote and approved the final version of the manuscript.}
\thanks{Supported by the National Library of Medicine of the National Institutes of Health (NIH) under Award Number DP2LM014268. The authors would like to thank Tori LeVier for her support in recruiting and managing participants. The content is solely the responsibility of the authors and does not necessarily represent the official views of the NIH. }
}

\maketitle

\begin{abstract}

Human-in-the-loop optimization (HILO) is a promising approach for personalizing visual prostheses by iteratively refining stimulus parameters based on user feedback. 
Previous work demonstrated HILO's efficacy in simulation, but its performance with human participants remains untested.
Here we evaluate HILO using sighted participants viewing simulated prosthetic vision to assess its ability to optimize stimulation strategies under realistic conditions. 
Participants selected between phosphenes generated by competing encoders to iteratively refine a deep stimulus encoder (DSE). 
We tested HILO in three conditions: standard optimization, threshold misspecifications, and out-of-distribution parameter sampling.
Participants consistently preferred HILO-generated stimuli over both a naïve encoder and the DSE alone, with log odds favoring HILO across all conditions.
We also observed key differences between human and simulated decision-making, highlighting the importance of validating optimization strategies with human participants. 
These findings support HILO as a viable approach for adapting visual prostheses to individuals.

\end{abstract}

\vspace{1ex}
\small
\textit{\textbf{Clinical Relevance---}}\textbf{Validating HILO with sighted participants viewing simulated prosthetic vision is an important step toward personalized calibration of future visual prostheses.}

\section{Introduction}

Visual prostheses are being developed to restore vision for individuals with incurable blindness by electrically stimulating functional cells along the visual pathway \cite{weiland_electrical_2016, fernandez_development_2018}.
Retinal \cite{luo_argusr_2016, stingl_interim_2017, palanker2022simultaneous} and cortical \cite{fernandez_visual_2021, barry2023preliminary, beauchamp2020dynamic} prostheses 
have enabled tasks such as object localization and supported mobility. 
However, the quality of vision provided by these implants remains limited. 
Phosphenes---the artificial visual percepts evoked by stimulation---vary widely across individuals \cite{beyeler_model_2019,sinclair_appearance_2016,fernandez_visual_2021, erickson-davis_what_2021} and often do not combine linearly \cite{hou_axonal_2024,barry_video-mode_2020,christie_sequential_2022}, suggesting the presence of a nonlinear transfer function between electrical stimuli and perceptual outcomes.

\begin{figure*}[!ht]
    \centering
    \includegraphics[width=\linewidth]{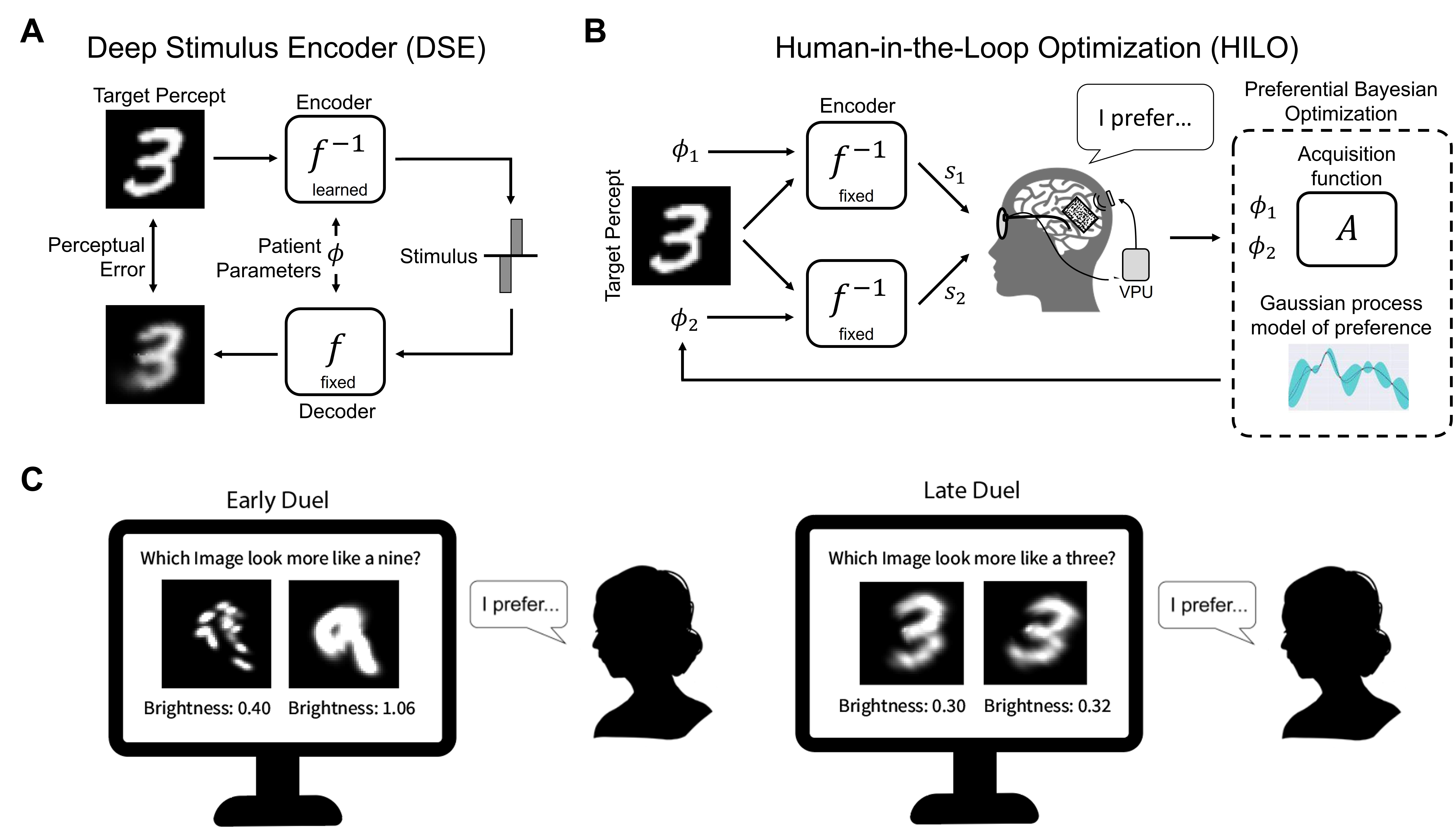}
    \caption{ \textbf{A)} Deep stimulus encoder (DSE). A forward model ($f$) predicts the perceptual response to visual stimuli based on user-specific parameters ($\phi$), while an encoder ($f^{-1}$) learns to minimize the perceptual error between predicted and target percepts. \textbf{B)} Human-in-the-loop optimization (HILO). The parameters from the DSE are refined using user preferences, collected through 60 binary comparison trials per condition. New parameter pairs are adaptively selected to efficiently converge on the most preferred percept. The target percept changes each iteration. Adapted under CC-BY from \protect\url{https://doi.org/10.48550/arXiv.2306.13104}. \textbf{C)} Example duels used to infer user preferences are presented to sighted participants on a computer monitor. Participants selected the preferred stimulus based on both shape and brightness.}
    \label{fig:overview}
\end{figure*}

To address these nonlinearities, computational models have been developed to predict perceptual responses to electrical stimulation. 
These \emph{forward} models use user-specific parameters to capture how stimulus properties affect the brightness and shape of phosphenes, as well as their interaction across electrodes \cite{granley_computational_2021, beyeler_model_2019, fine_virtual_2024, van_der_grinten_towards_2024, granley2022adapting}. 
While forward models enable a mechanistic understanding of prosthetic vision, their utility in real-world applications hinges on the ability to invert them: that is, to determine the optimal stimulus parameters required to elicit a desired percept. 
\Acp{DSE} have been proposed for this purpose (Fig.~\ref{fig:overview}A), leveraging deep learning to approximate the \emph{inverse} mapping from percepts to stimuli \cite{de_ruyter_van_steveninck_end--end_2022, relic_deep_2022, granley_hybrid_2022, van_der_grinten_towards_2024, kuccukouglu2025end}. 
However, \acp{DSE} require precise knowledge of user-specific parameters, which is often infeasible due to limited data availability or the inherent variability across users~\cite{granley_human---loop_2023}.
Moreover, interviews with blind prosthesis users have highlighted gaps between laboratory models and real-world experiences, underscoring the importance of designing technologies that adapt to individual needs and usability constraints~\cite{nadolskis_aligning_2024}.

To optimize stimulus strategies from limited human feedback, the \ac{HILO} framework was introduced to the field by Fauvel and Chalk~\cite{fauvel_human---loop_2022} (Fig.~\ref{fig:overview}B). 
Their work applied \ac{HILO} to a linear phosphene model, showing that parameter estimation was feasible using pairwise comparisons of simulated percepts (\emph{duels}). 
Granley \textit{et al.}~\cite{granley_human---loop_2023} extended this approach by incorporating \acp{DSE} and testing it under more realistic conditions, including model misspecifications and noisy user feedback. 
Their results suggested that \ac{HILO} could effectively personalize stimulus encoding, even when simulated users selected randomly in two out of three trials. 
However, these simulations relied on predefined probabilistic rules, where choices were more likely when a percept had a lower computed error metric. 
Human decision making, by contrast, is inherently more variable and may not align with model assumptions, leaving open the question of whether \ac{HILO} would perform similarly when real users provide feedback.

In this study, we take an intermediate step toward real-world validation by incorporating sighted participants viewing \ac{SPV} stimuli on a monitor (Fig.~\ref{fig:overview}C). This approach enables systematic testing of whether \ac{HILO} can adapt to individual perceptual variability, maintain robustness to model misspecifications, and remain practical under real-world decision-making constraints. Importantly, we evaluated HILO not only under standard conditions but also in out-of-distribution scenarios, testing its generalizability beyond the training range. We also systematically varied key phosphene properties, such as size and elongation, highlighting how perceptual distortions affect optimization performance and motivating future efforts to improve phosphene focality. By replacing simulated user decisions with human feedback, this study provides critical evidence for the utility of HILO in personalizing neuroprosthetic devices.


\section{Methods}

\subsection{Participants}

Seventeen sighted undergraduate students from the University of California, Santa Barbara, participated in the study (10 female, 7 male; ages 18--21; $M = 19.6$, $SD = 1.12$).
Participants were recruited through university-wide announcements and provided informed consent prior to participation. 
All participants reported normal or corrected-to-normal vision and no history of neurological or visual impairments.
Participants were briefed on the study's purpose and tasks before beginning the experiment.
The study was approved by the Institutional Review Board at the University of California, Santa Barbara.

\subsection{Task}

Participants viewed simulated prosthetic vision on a monitor and completed a series of trials aimed at optimizing the mapping between simulated electrical stimuli and perceived visual representations. 
Each trial, referred to as a ``duel," presented two full percepts generated for a target image. 
Participants selected the percept they judgedt to better match the target, which was described textually (e.g., ``number eight") to encourage independent mental representations.
To prevent selection bias, the positions of the two stimuli (left or right) were randomized across trials.

\begin{figure}[!bh]
    \centering
    \includegraphics[width=\linewidth]{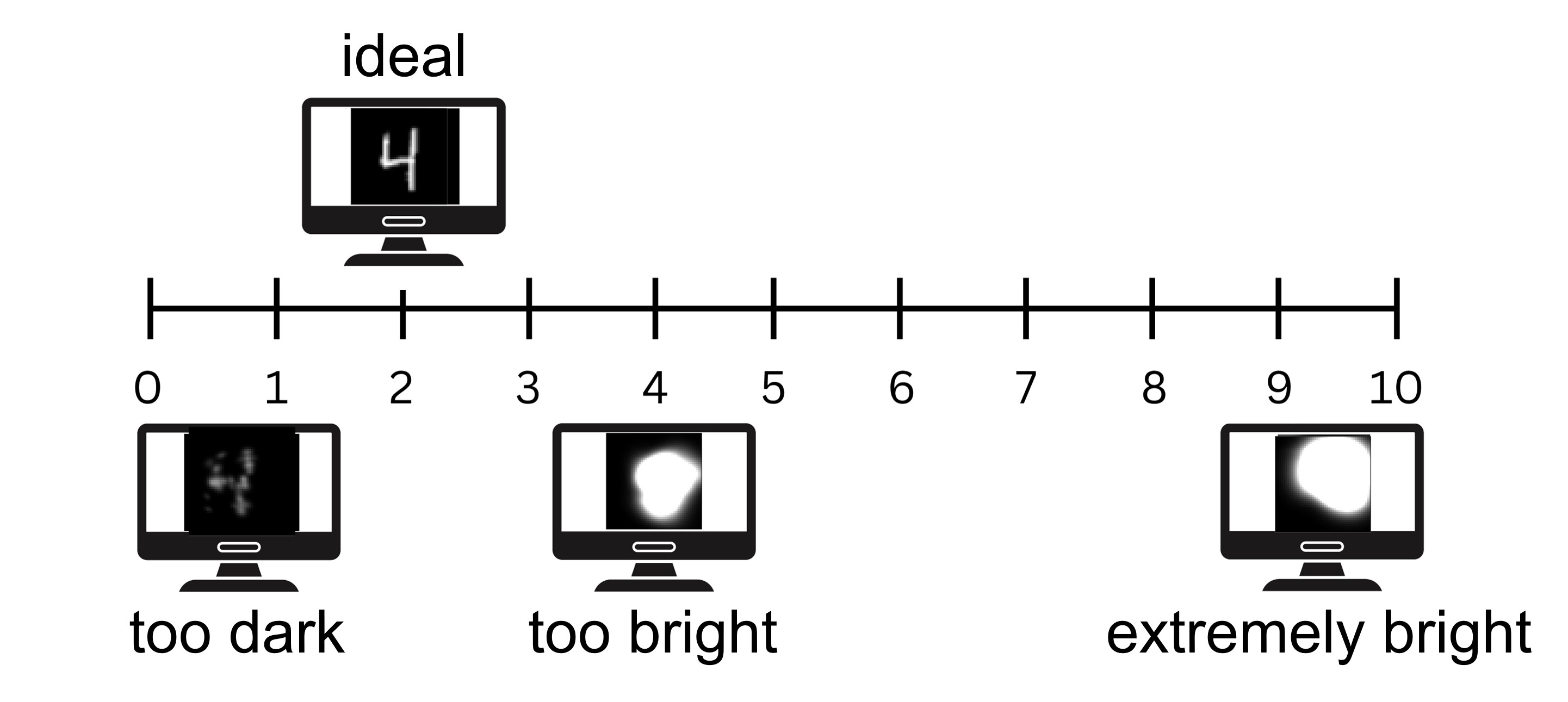}
    \caption{Example stimuli illustrating the range of phosphene brightness levels shown to participants. A value of 0 represents complete darkness, 2 is the ideal brightness for a retinal prosthesis user, 5 is overly bright, and 10 is extremely bright, with white filling most of the stimulus area.}
    \label{fig:stimuli-brightness}
\end{figure}

In practice, electrical stimulation can evoke phosphenes that are excessively bright or even painful \cite{Brindley1968, Dobelle1974, erickson-davis_what_2021, luo_argusr_2016}. To account for the limited dynamic range of computer monitors, numerical brightness values were displayed alongside each stimulus. During the tutorial phase, participants were introduced to the brightness scale: 1 represented threshold brightness (too dim), 2 was twice the threshold (ideal), and higher values indicated increasing excessive brightness (Fig.~\ref{fig:stimuli-brightness}).

The experiment consisted of three phases:
\begin{itemize}
    \item \emph{Tutorial phase:} Participants completed practice trials to familiarize themselves with the task, including how to make selections and interpret brightness values.
    \item \emph{Optimization phase:} Participants completed 60 duels, during which the \ac{HILO} framework iteratively refined the simulated user parameters $\phi$ based on their preferences.
    \item \emph{Evaluation phase:} Participants completed 39 additional duels, comparing the optimized \ac{DSE} against a naïve baseline encoder.
\end{itemize}

\subsection{Simulated Prosthetic Vision}

We employed the experimentally validated computational model from \cite{granley_human---loop_2023} to replicate the perceptual experience of an epiretinal implant.
This model represents phosphenes as multivariate Gaussian blobs,
whose size, shape, and brightness are determined by both electrode location and stimulus properties such as amplitude, frequency, and pulse duration~\cite{nanduri_frequency_2012, horsager_predicting_2009, beyeler_model_2019}. 
Importantly, the appearance of these phosphenes varies across individuals due to differences in retinal anatomy and neural processing.

To capture this variability, the model includes a set of 13 user-specific parameters, $\phi$, which govern key aspects of phosphene appearance.
Among these, $\rho$ (in microns) controls overall phosphene size, while $\lambda$ (ranging from 0 for circular phosphenes to near 1 for highly elongated ones) modulates elongation along the trajectory of underlying axon pathways.
Other parameters influence brightness scaling, the spread of axonal streaking effects, the location of the optic disc (which influences axonal trajectories), the implant location and orientation relative to the fovea, and electrode sensitivity. For full details, see~\cite{granley_human---loop_2023}.
Each participant was randomly assigned a unique set of $\phi$ values, simulating the perceptual variability observed in real prosthesis users.

The model maps an electrical stimulus $\mathbf{s} \in \mathbb{R}^{n_e \times 3}$ to a visual percept. 
Each phosphene is modeled as a Gaussian blob with center $\mu_e$, covariance matrix $\Sigma_e$, and brightness $b_e$, determined by both $\phi$ and the applied stimulus~\cite{granley_human---loop_2023}:
\begin{equation}
    b(x, y) = 2\pi b_e\det{(\mathbf{\Sigma_e}})\,\, \mathcal{N}([x, y]^\top
| \mu_e, \mathbf{\Sigma_e}),
\end{equation}
where $\mathcal{N}$ represents a Gaussian distribution, $\det{(\Sigma_e)}$ ensures proper normalization of the percept brightness, and $\Sigma_e$ encodes phosphene shape and orientation, incorporating effects of retinal fiber structure.

To form the final percept, phosphenes from all stimulated electrodes are summed across the visual field.
This summation introduces nonlinear interactions between adjacent phosphenes, which can lead to perceptual distortions similar to those reported by real prosthesis users~\cite{granley_computational_2021}.

This model serves as the foundation for generating the \ac{SPV} stimuli used in this study.

\subsection{Human-in-the-Loop Optimization (HILO)}

Following \cite{granley_human---loop_2023}, a neural network \ac{DSE} was trained to invert the forward model, generating stimulus parameters that elicit a phosphene most closely matching a target image. 
The \ac{DSE} was designed as a fully connected feedforward network with multiple residual blocks, each containing batch normalization and leaky ReLU activations to stabilize training and improve generalization.
The model takes as input the target image and the user-specific parameters $\phi$, which encode individual variations in phosphene perception.
By training the \ac{DSE} across a broad distribution of $\phi$ values~\cite{granley_human---loop_2023}, it learns to produce optimized stimuli tailored to diverse user profiles.

To further refine these user-specific parameters, we employed human-in-the-loop optimization (\ac{HILO}), iteratively updating $\phi$ estimates based on pairwise comparisons provided by human participants \cite{granley_human---loop_2023}.
Preferences between two candidate parameter sets, $\phi_1$ and $\phi_2$, were modeled using a Gaussian process:
\begin{equation}
P(\phi_1 \succ \phi_2|g)= \Phi \big( g(\phi_1)  - g(\phi_2) \big),
\end{equation}
where $\Phi$ was the normal cumulative distribution, and $g(\phi_i)$ represented the preference function learned by the Gaussian process.
A larger value of $g(\phi_1)$ relative to $g(\phi_2)$ indicated a higher likelihood that the participant preferred $\phi_1$.

We employed the Maximally Uncertain Challenge acquisition function \cite{fauvel_human---loop_2022} to balance exploration (identifying uncertain parameter regions) and exploitation (refining the best-known parameters).
Specifically:
\begin{align}
    \phi_1 & \rightarrow \underset{\phi}{\arg \max} \ \mathbb{E}_{p(g|\mathscr{D})}[g(\phi)], \\
    \phi_2 & \rightarrow \underset{\phi}{\arg \max} \  \mathbb{V}_{p(g|\mathscr{D})}[\Phi(g(\phi)-g(\phi_1))],
\end{align}
where $\phi_1$ was the ``champion" (the current best parameter set), and $\phi_2$ was the ``challenger" (the parameter set for which preferences were most uncertain). 
Here, $\mathbb{E}$ and $\mathbb{V}$ denote the expectation and variance, respectively.

Participants guided the optimization process by viewing percepts generated for a target MNIST digit encoded with $\phi_1$ and $\phi_2$. 
They selected the percept they perceived as more similar to the target. 
These pairwise preferences were then used to update the Gaussian process model, iteratively refining the estimates of the participant’s optimal parameters $\phi$.

\begin{figure*}[t!]
    \centering
 \includegraphics[width=1\linewidth]{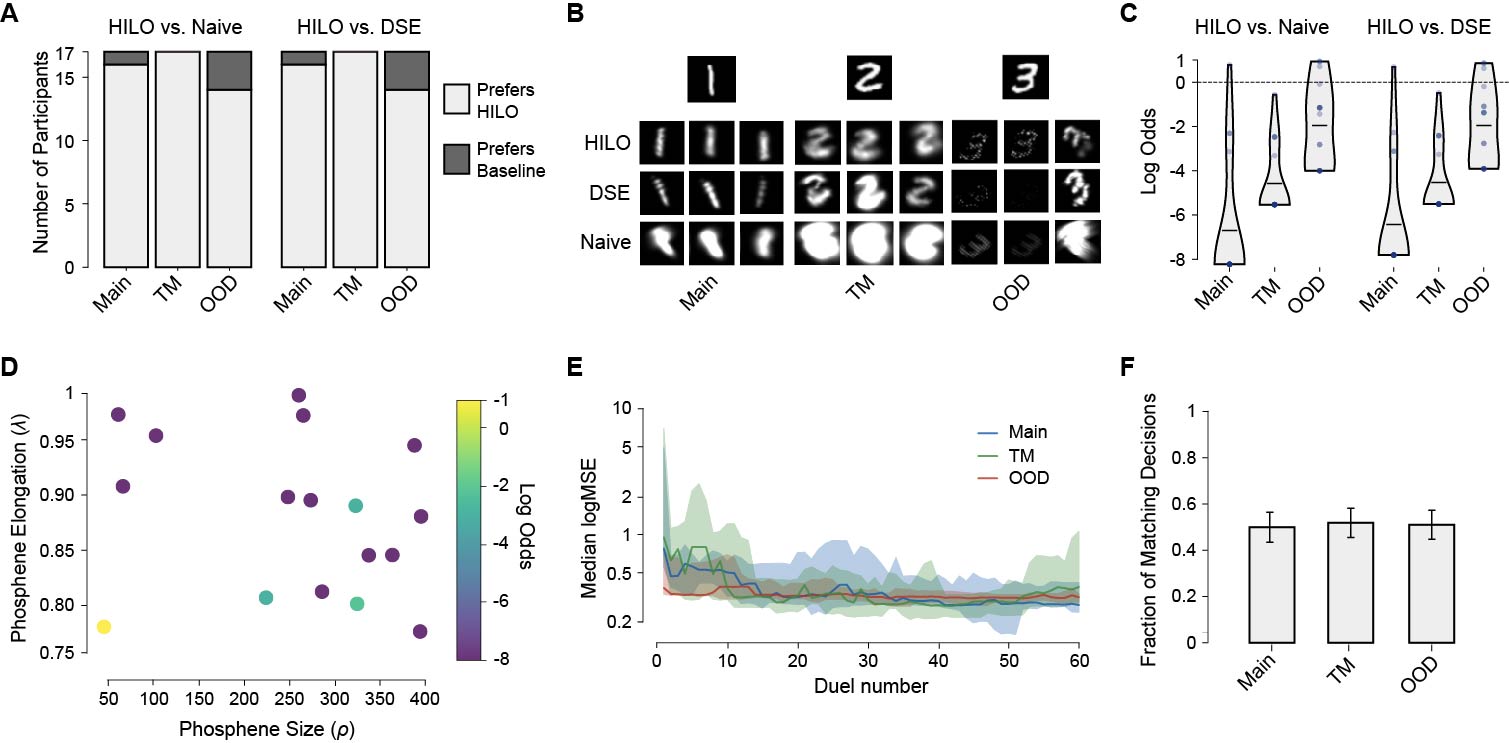}
    \caption{\textbf{A)} Number of participants who significantly preferred \acf{HILO} over the naïve encoder (left) and the \acf{DSE} without HILO (right), based on log odds less than 0 in a linear mixed-effects model. \textbf{B)} Example percepts generated by the \ac{HILO} encoder, the \ac{DSE} without HILO, and the naïve encoder for three participants across the three experimental conditions. \textbf{C)} Distribution of log odds for \ac{HILO} across the three experiments. \textbf{D)} Distribution of the two main user-specific parameters, $\rho$ (phosphene size) and $\lambda$ (axon-aligned elongation), colored by the log odds indicating preference for \ac{HILO} in the main experiment. \textbf{E)} Median \acf{MSE} over the course of optimization for each experiment, with shaded regions denoting the interquartile range (IQR). \textbf{F)} Proportion of duels where participant decisions matched those of the simulated agent.}
    \label{fig:results-main}
\end{figure*}

\subsection{Experimental Conditions}

The study included three experimental conditions, designed to evaluate the performance and robustness of the \ac{HILO} framework under varying levels of complexity and realism:
\begin{itemize}
    \item \emph{Main experiment:} This was the standard experiment adapted from \cite{granley_human---loop_2023}, where the unknown parameters $\phi$ for simulated users were iteratively optimized using participant feedback.
    \item \emph{Threshold misspecification (TM):} This experiment introduced errors in the assumed threshold amplitudes of up to 300\%, testing the framework's robustness to inaccuracies in user-specific parameters.
    \item \emph{Out-of-distribution (OOD):} In this experiment, the ground-truth $\phi$ values for simulated users were drawn from a distribution outside the training range of the deep stimulus encoder (\ac{DSE}), evaluating the framework's ability to generalize to unseen parameter configurations.
\end{itemize}

In each condition, the \ac{HILO} framework was evaluated against two baseline models:
\begin{itemize}
    \item \emph{Naïve encoder:} The approach traditionally used by prostheses, where the target image is reduced to the electrode array resolution, and each pixel’s grayscale value is directly scaled to stimulus amplitude.
    \item \emph{\Acf{DSE}:} The same deep stimulus encoder as used in \ac{HILO}, but instead of user-specific parameters being tuned, they are guessed as the mean of the observed ranges reported in ~\cite{granley_human---loop_2023}.
\end{itemize}

\section{Results}

\subsection{Participant Preferences for HILO vs. Other Encoders}

Across all three experimental conditions, participants consistently preferred \ac{HILO}-optimized stimuli over both the naïve encoder and the non-personalized \ac{DSE} (Fig.~\ref{fig:results-main}A). 
Example percepts across the three experimental conditions are shown in Fig.~\ref{fig:results-main}B.
Statistical analysis was performed with a linear mixed-effects logistic regression model, with population and per-subject effects.
Across subjects, \ac{HILO} was significantly preferred over the naïve and \ac{DSE} encoders for all three experiments ($p<0.0001$).
In the main experiment, 16 out of 17 participants favored \ac{HILO} over the other encoders (defined as having a subject-level log odds of less than 0, indicating preference for \ac{HILO}). 
In the threshold misspecification condition, all participants selected \ac{HILO}. 
In the out-of-distribution condition, 14 out of 17 participants preferred \ac{HILO}.

Percepts generated by \ac{HILO} generally retained greater structure and recognizability compared to baseline methods. 
In the main experiment, both \ac{HILO} and \ac{DSE} produced structured percepts that aligned with target images, whereas the naïve encoder often resulted in highly distorted or unrecognizable shapes. 
Under threshold misspecification, \ac{HILO} maintained consistent percepts, whereas \ac{DSE} outputs varied widely, sometimes producing overexposed phosphenes. 
In the OOD condition, distortions appeared across all methods, though \ac{HILO} percepts remained more structured and interpretable than those generated by the alternatives.

The distributions of log odds per participant are shown in Fig.~\ref{fig:results-main}C. 
The mean log odds for HILO over the naïve encoder were $-6.70 \pm  0.71$ (main experiment), $-4.57 \pm  0.39$ (threshold misspecification), and $-1.95 \pm 0.46$ (OOD condition). 
Lower log odds indicate stronger preference for \ac{HILO}, confirming its advantage across conditions. 
Although more participants chose \ac{HILO} in the threshold misspecification experiment than in the main experiment, their preference was weaker, as reflected in the less negative log odds.

\subsection{Log Odds Across Individual User-Specific Parameters}

To assess how individual user-specific factors influenced preference trends, log odds were examined as a function of $\rho$ (phosphene size) and $\lambda$ (phosphene elongation) \cite{beyeler_model_2019}
Fig.~\ref{fig:results-main}D.
Participants who were assigned larger, more elongated phosphenes exhibited higher log odds in favor of HILO, while those with smaller, minimally distorted percepts showed lower log odds differences between encoders. 
This pattern suggests that HILO provides the greatest benefit when percepts are highly distorted but offers less improvement when phosphenes are already relatively structured.

\subsection{Comparison Between Human and Simulated Preferences}

\Ac{MSE} decreased over the course of optimization (Fig.~\ref{fig:results-main}E), confirming that participant choices contributed to improved stimulus encoding. 
The final loss in the main experiment converged to 0.27, with similar performance in the other two conditions.
In contrast, purely simulated experiments from Granley et al.\cite{granley_human---loop_2023}, where the goal was to directly minimize \ac{MSE}, achieved a much lower final loss of 0.07. 
The slower rate of loss reduction in human trials suggests that participants were not strictly optimizing for pixel-wise reconstruction accuracy, instead incorporating additional perceptual or cognitive factors into their choices.

To further quantify these differences, we compared participant selections to those predicted by the simulated agent from Granley et al.\cite{granley_human---loop_2023} (Fig.\ref{fig:results-main}F).
Across all three experiments, participants made the same choice as the simulated agent in only about 50\% of trials, indicating that human decision-making diverges from the assumptions made in purely simulation-based studies. 
Despite this variability, participants still overwhelmingly preferred \ac{HILO}, demonstrating its robustness to real-world human responses and reinforcing the need for empirical validation beyond theoretical models.

\section{Discussion}

This study demonstrates that \acf{HILO} successfully personalizes \acfp{DSE} for individual users, even in the presence of model inaccuracies and noisy human feedback. 
Sighted participants viewing simulated prosthetic vision consistently preferred \ac{HILO}-optimized stimuli over both naïve encoding and a standard \ac{DSE}, reinforcing the viability of this approach for adapting neuroprosthetic devices to individual users. 
These findings provide critical evidence supporting the application of \ac{HILO} to real-world neuroprosthetic calibration.

\subsection{HILO Stimuli Are Preferred Across Conditions}

Across all experimental conditions, participants consistently favored \ac{HILO}-optimized stimuli over baseline encoders. Even when model parameters were intentionally misspecified, \ac{HILO} maintained its advantage, demonstrating robustness to variations in user-specific parameters. These results suggest that, even when the underlying model is imperfect, \ac{HILO} can continue to improve performance by optimizing toward patient preferences, which remain informative despite deviations from model assumptions.

In the out-of-distribution condition, where the \ac{DSE} was not trained on the tested parameter ranges, a small subset of participants preferred the baseline encoder.
This suggests that HILO is most effective when operating within a parameter space it has encountered during training, and its generalization outside of this range may require additional refinements. 

Nonetheless, log odds analyses confirmed that \ac{HILO} still provided a perceptual advantage for most participants, reinforcing its adaptability to individual differences in perception.

\subsection{Human Decision-Making Differs from Simulated Users}

Previous studies evaluating \ac{HILO}~\cite{fauvel_human---loop_2022, granley_human---loop_2023} relied entirely on simulated users, whose decisions followed predefined loss functions. 
While Granley et al.~\cite{granley_human---loop_2023} attempted to account for human variability by introducing artificial noise (i.e., demonstrating that \ac{HILO} remained effective even when simulated users made random choices in two out of three trials), real human decision-making may not align with these assumptions.

Our results confirm that human choices diverged from simulated predictions. 
Although \ac{MSE} decreased over the course of optimization, the final loss was significantly higher than in purely simulated experiments~\cite{granley_human---loop_2023}, suggesting that human participants did not strictly optimize for pixel-wise accuracy.
Additionally, participants selected the same choices as the simulated agent in only about 50\% of trials.
This suggests that human decision-making incorporates perceptual factors beyond those captured by the computational model. 

Despite this divergence, participants still overwhelmingly preferred \ac{HILO}, reinforcing the need for validation with real users rather than relying solely on theoretical models.

\subsection{Limitations and Future Work}

While this study provides strong support for the use of \ac{HILO} in personalizing neuroprosthetic devices, several important challenges remain. The next step is testing \ac{HILO} with blind prosthesis users, though this presents practical and safety considerations. Deep learning models can produce unpredictable outputs, and while most neuroprosthetic devices include firmware safeguards to enforce stimulation limits, additional validation will be needed to ensure that optimized stimuli do not generate unintended or adverse effects.

Additional limitations of the present study include the use of a static computer screen to display simulated percepts, which does not fully capture the dynamic experience of artificial vision in everyday life. Moreover, the simulated percepts shown to participants did not incorporate the temporal fading that real prosthesis users experience, potentially affecting the ecological validity of participant responses.

Future work should refine the optimization objective to better align with human perceptual judgments, moving beyond pixel-wise error metrics toward loss functions that capture higher-level visual features~\cite{granley_hybrid_2022,prashnani_pieapp_2018}. Extending validation to blind participants and testing under more dynamic, real-world conditions will be critical for clinical translation. Recent work underscores the need to align prosthetic development with user needs~\cite{nadolskis_aligning_2024} and the realities of assistive technology use in daily life~\cite{turkstra_assistive_2025}, emphasizing that future optimization must prioritize not only perceptual similarity but also usability and functional task performance. Additionally, optimizing stimulation to align with latent neural representations in the visual cortex~\cite{granley_beyond_2024} offers a promising path toward deeper integration between artificial stimulation and biological processing. Together, these directions move beyond technical optimization toward meaningful, user-centered neuroprosthetic design.

Beyond visual prostheses, the HILO framework may generalize to other sensory neuroprosthetic systems, such as cochlear implants and tactile feedback devices, where forward models and deep encoders are increasingly being adopted~\cite{Okorokova,Dorman,drakopoulos_dnn-based_2023,lesica_harnessing_2021}. Recent work on brain co-processors~\cite{rao_brain_2023} emphasizes the growing importance of closed-loop systems that jointly optimize neural decoding and stimulation in collaboration with the brain itself. Our findings support this broader vision by demonstrating that real human feedback can successfully guide the adaptation of stimulus encoding strategies, advancing toward tighter integration of artificial and biological neural systems. By establishing the viability of HILO with human participants, this study represents an important step toward fully individualized optimization of neuroprosthetic technologies.

\bibliographystyle{IEEEtran}
\bibliography{2025-EMBC-HILO.bib}

\begin{thebibliography}{10}
\providecommand{\url}[1]{#1}
\csname url@samestyle\endcsname
\providecommand{\newblock}{\relax}
\providecommand{\bibinfo}[2]{#2}
\providecommand{\BIBentrySTDinterwordspacing}{\spaceskip=0pt\relax}
\providecommand{\BIBentryALTinterwordstretchfactor}{4}
\providecommand{\BIBentryALTinterwordspacing}{\spaceskip=\fontdimen2\font plus
\BIBentryALTinterwordstretchfactor\fontdimen3\font minus \fontdimen4\font\relax}
\providecommand{\BIBforeignlanguage}[2]{{%
\expandafter\ifx\csname l@#1\endcsname\relax
\typeout{** WARNING: IEEEtran.bst: No hyphenation pattern has been}%
\typeout{** loaded for the language `#1'. Using the pattern for}%
\typeout{** the default language instead.}%
\else
\language=\csname l@#1\endcsname
\fi
#2}}
\providecommand{\BIBdecl}{\relax}
\BIBdecl

\bibitem{weiland_electrical_2016}
\BIBentryALTinterwordspacing
J.~D. Weiland, S.~T. Walston, and M.~S. Humayun, ``Electrical {Stimulation} of the {Retina} to {Produce} {Artificial} {Vision},'' \emph{Annual Review of Vision Science}, vol.~2, no.~1, pp. 273--294, 2016. [Online]. Available: \url{https://doi.org/10.1146/annurev-vision-111815-114425}
\BIBentrySTDinterwordspacing

\bibitem{fernandez_development_2018}
\BIBentryALTinterwordspacing
E.~Fernandez, ``Development of visual {Neuroprostheses}: trends and challenges,'' \emph{Bioelectronic Medicine}, vol.~4, no.~1, p.~12, Aug. 2018. [Online]. Available: \url{https://doi.org/10.1186/s42234-018-0013-8}
\BIBentrySTDinterwordspacing

\bibitem{luo_argusr_2016}
Y.~H. Luo and L.~da~Cruz, ``The {Argus}(({R})) {II} {Retinal} {Prosthesis} {System},'' \emph{Prog Retin Eye Res}, vol.~50, pp. 89--107, Jan. 2016.

\bibitem{stingl_interim_2017}
\BIBentryALTinterwordspacing
K.~Stingl, R.~Schippert, K.~U. Bartz-Schmidt, D.~Besch, C.~L. Cottriall, T.~L. Edwards, F.~Gekeler, U.~Greppmaier, K.~Kiel, A.~Koitschev, L.~Kühlewein, R.~E. MacLaren, J.~D. Ramsden, J.~Roider, A.~Rothermel, H.~Sachs, G.~S. Schröder, J.~Tode, N.~Troelenberg, and E.~Zrenner, ``\BIBforeignlanguage{English}{Interim {Results} of a {Multicenter} {Trial} with the {New} {Electronic} {Subretinal} {Implant} {Alpha} {AMS} in 15 {Patients} {Blind} from {Inherited} {Retinal} {Degenerations}},'' \emph{\BIBforeignlanguage{English}{Frontiers in Neuroscience}}, vol.~11, 2017, publisher: Frontiers. [Online]. Available: \url{https://www.frontiersin.org/articles/10.3389/fnins.2017.00445/full}
\BIBentrySTDinterwordspacing

\bibitem{palanker2022simultaneous}
D.~Palanker, Y.~Le~Mer, S.~Mohand-Said, and J.-A. Sahel, ``{Simultaneous perception of prosthetic and natural vision in AMD patients},'' \emph{Nature communications}, vol.~13, no.~1, p. 513, 2022.

\bibitem{fernandez_visual_2021}
E.~Fernández, A.~Alfaro, C.~Soto-Sánchez, P.~Gonzalez-Lopez, A.~M. Lozano, S.~Peña, M.~D. Grima, A.~Rodil, B.~Gómez, X.~Chen, P.~R. Roelfsema, J.~D. Rolston, T.~S. Davis, and R.~A. Normann, ``\BIBforeignlanguage{eng}{Visual percepts evoked with an intracortical 96-channel microelectrode array inserted in human occipital cortex},'' \emph{\BIBforeignlanguage{eng}{The Journal of Clinical Investigation}}, vol. 131, no.~23, p. e151331, Dec. 2021.

\bibitem{barry2023preliminary}
M.~P. Barry, R.~Sadeghi, V.~L. Towle, K.~Stipp, H.~Puhov, W.~Diaz, P.~Grant, F.~T. Collison, F.~J. Lane, J.~P. Szlyk \emph{et~al.}, ``{Preliminary visual function for the first human with the Intracortical Visual Prosthesis (ICVP)},'' \emph{Investigative Ophthalmology \& Visual Science}, vol.~64, no.~8, pp. 2842--2842, 2023.

\bibitem{beauchamp2020dynamic}
M.~S. Beauchamp, D.~Oswalt, P.~Sun, B.~L. Foster, J.~F. Magnotti, S.~Niketeghad, N.~Pouratian, W.~H. Bosking, and D.~Yoshor, ``Dynamic stimulation of visual cortex produces form vision in sighted and blind humans,'' \emph{Cell}, vol. 181, no.~4, pp. 774--783, 2020.

\bibitem{beyeler_model_2019}
\BIBentryALTinterwordspacing
M.~Beyeler, D.~Nanduri, J.~D. Weiland, A.~Rokem, G.~M. Boynton, and I.~Fine, ``\BIBforeignlanguage{en}{A model of ganglion axon pathways accounts for percepts elicited by retinal implants},'' \emph{\BIBforeignlanguage{en}{Scientific Reports}}, vol.~9, no.~1, pp. 1--16, Jun. 2019. [Online]. Available: \url{https://www.nature.com/articles/s41598-019-45416-4}
\BIBentrySTDinterwordspacing

\bibitem{sinclair_appearance_2016}
\BIBentryALTinterwordspacing
N.~C. Sinclair, M.~N. Shivdasani, T.~Perera, L.~N. Gillespie, H.~J. McDermott, L.~N. Ayton, and P.~J. Blamey, ``\BIBforeignlanguage{en}{The {Appearance} of {Phosphenes} {Elicited} {Using} a {Suprachoroidal} {Retinal} {Prosthesis}},'' \emph{\BIBforeignlanguage{en}{Investigative Ophthalmology \& Visual Science}}, vol.~57, no.~11, pp. 4948--4961, Sep. 2016. [Online]. Available: \url{https://iovs.arvojournals.org/article.aspx?articleid=2556019}
\BIBentrySTDinterwordspacing

\bibitem{erickson-davis_what_2021}
\BIBentryALTinterwordspacing
C.~Erickson-Davis and H.~Korzybska, ``\BIBforeignlanguage{en}{What do blind people “see” with retinal prostheses? {Observations} and qualitative reports of epiretinal implant users},'' \emph{\BIBforeignlanguage{en}{PLOS ONE}}, vol.~16, no.~2, p. e0229189, Feb. 2021, publisher: Public Library of Science. [Online]. Available: \url{https://journals.plos.org/plosone/article?id=10.1371/journal.pone.0229189}
\BIBentrySTDinterwordspacing

\bibitem{hou_axonal_2024}
\BIBentryALTinterwordspacing
Y.~Hou, D.~Nanduri, J.~Granley, J.~D. Weiland, and M.~Beyeler, ``\BIBforeignlanguage{en}{Axonal stimulation affects the linear summation of single-point perception in three {Argus} {II} users},'' \emph{\BIBforeignlanguage{en}{Journal of Neural Engineering}}, vol.~21, no.~2, p. 026031, Apr. 2024, publisher: IOP Publishing. [Online]. Available: \url{https://dx.doi.org/10.1088/1741-2552/ad31c4}
\BIBentrySTDinterwordspacing

\bibitem{barry_video-mode_2020}
M.~P. Barry, M.~Armenta~Salas, U.~Patel, V.~Wuyyuru, S.~Niketeghad, W.~H. Bosking, D.~Yoshor, J.~D. Dorn, and N.~Pouratian, ``Video-mode percepts are smaller than sums of single-electrode phosphenes with the {Orion}® visual cortical prosthesis,'' \emph{Investigative Ophthalmology \& Visual Science}, vol.~61, no.~7, p. 927, Jun. 2020.

\bibitem{christie_sequential_2022}
\BIBentryALTinterwordspacing
B.~Christie, R.~Sadeghi, A.~Kartha, A.~Caspi, F.~V. Tenore, R.~L. Klatzky, G.~Dagnelie, and S.~Billings, ``\BIBforeignlanguage{en}{Sequential epiretinal stimulation improves discrimination in simple shape discrimination tasks only},'' \emph{\BIBforeignlanguage{en}{Journal of Neural Engineering}}, vol.~19, no.~3, p. 036033, Jun. 2022, publisher: IOP Publishing. [Online]. Available: \url{https://dx.doi.org/10.1088/1741-2552/ac7326}
\BIBentrySTDinterwordspacing

\bibitem{granley_computational_2021}
\BIBentryALTinterwordspacing
J.~Granley and M.~Beyeler, ``A {Computational} {Model} of {Phosphene} {Appearance} for {Epiretinal} {Prostheses},'' \emph{Annual International Conference of the IEEE Engineering in Medicine and Biology Society. IEEE Engineering in Medicine and Biology Society. Annual International Conference}, vol. 2021, pp. 4477--4481, Nov. 2021. [Online]. Available: \url{https://www.ncbi.nlm.nih.gov/pmc/articles/PMC9255280/}
\BIBentrySTDinterwordspacing

\bibitem{fine_virtual_2024}
\BIBentryALTinterwordspacing
I.~Fine and G.~M. Boynton, ``\BIBforeignlanguage{en}{A virtual patient simulation modeling the neural and perceptual effects of human visual cortical stimulation, from pulse trains to percepts},'' \emph{\BIBforeignlanguage{en}{Scientific Reports}}, vol.~14, no.~1, p. 17400, Jul. 2024, publisher: Nature Publishing Group. [Online]. Available: \url{https://www.nature.com/articles/s41598-024-65337-1}
\BIBentrySTDinterwordspacing

\bibitem{van_der_grinten_towards_2024}
\BIBentryALTinterwordspacing
M.~van~der Grinten, J.~de~Ruyter~van Steveninck, A.~Lozano, L.~Pijnacker, B.~Rueckauer, P.~Roelfsema, M.~van Gerven, R.~van Wezel, U.~Güçlü, and Y.~Güçlütürk, ``Towards biologically plausible phosphene simulation for the differentiable optimization of visual cortical prostheses,'' \emph{eLife}, vol.~13, p. e85812, Feb. 2024, publisher: eLife Sciences Publications, Ltd. [Online]. Available: \url{https://doi.org/10.7554/eLife.85812}
\BIBentrySTDinterwordspacing

\bibitem{granley2022adapting}
J.~Granley, A.~Riedel, and M.~Beyeler, ``Adapting brain-like neural networks for modeling cortical visual prostheses,'' \emph{arXiv preprint arXiv:2209.13561}, 2022.

\bibitem{de_ruyter_van_steveninck_end--end_2022}
\BIBentryALTinterwordspacing
J.~de~Ruyter~van Steveninck, U.~Güçlü, R.~van Wezel, and M.~van Gerven, ``End-to-end optimization of prosthetic vision,'' \emph{Journal of Vision}, vol.~22, no.~2, p.~20, Feb. 2022. [Online]. Available: \url{https://doi.org/10.1167/jov.22.2.20}
\BIBentrySTDinterwordspacing

\bibitem{relic_deep_2022}
\BIBentryALTinterwordspacing
L.~Relic, B.~Zhang, Y.-L. Tuan, and M.~Beyeler, ``Deep {Learning}\&\#x2013;{Based} {Perceptual} {Stimulus} {Encoder} for {Bionic} {Vision},'' in \emph{Augmented {Humans} 2022}, ser. {AHs} 2022.\hskip 1em plus 0.5em minus 0.4em\relax New York, NY, USA: Association for Computing Machinery, Mar. 2022, pp. 323--325. [Online]. Available: \url{https://doi.org/10.1145/3519391.3524034}
\BIBentrySTDinterwordspacing

\bibitem{granley_hybrid_2022}
\BIBentryALTinterwordspacing
J.~Granley, L.~Relic, and M.~Beyeler, ``\BIBforeignlanguage{en}{Hybrid {Neural} {Autoencoders} for {Stimulus} {Encoding} in {Visual} and {Other} {Sensory} {Neuroprostheses}},'' in \emph{\BIBforeignlanguage{en}{Advances in {Neural} {Information} {Processing} {Systems}}}, vol.~35, Dec. 2022, pp. 22\,671--22\,685. [Online]. Available: \url{https://papers.nips.cc/paper_files/paper/2022/hash/8e9a6582caa59fda0302349702965171-Abstract-Conference.html}
\BIBentrySTDinterwordspacing

\bibitem{kuccukouglu2025end}
B.~K{\"u}{\c{c}}{\"u}ko{\u{g}}lu, B.~Rueckauer, J.~de~Ruyter~van Steveninck, M.~van~der Grinten, Y.~G{\"u}{\c{c}}l{\"u}t{\"u}rk, P.~R. Roelfsema, U.~G{\"u}{\c{c}}l{\"u}, and M.~van Gerven, ``End-to-end learning of safe stimulation parameters for cortical neuroprosthetic vision,'' \emph{bioRxiv}, pp. 2025--01, 2025.

\bibitem{granley_human---loop_2023}
\BIBentryALTinterwordspacing
J.~Granley, T.~Fauvel, M.~Chalk, and M.~Beyeler, ``\BIBforeignlanguage{en}{Human-in-the-{Loop} {Optimization} for {Deep} {Stimulus} {Encoding} in {Visual} {Prostheses}},'' in \emph{\BIBforeignlanguage{en}{Advances in {Neural} {Information} {Processing} {Systems}}}, Nov. 2023. [Online]. Available: \url{https://openreview.net/forum?id=ZED5wdGous}
\BIBentrySTDinterwordspacing

\bibitem{nadolskis_aligning_2024}
\BIBentryALTinterwordspacing
L.~Nadolskis, L.~M. Turkstra, E.~Larnyo, and M.~Beyeler, ``Aligning {Visual} {Prosthetic} {Development} {With} {Implantee} {Needs},'' \emph{Translational Vision Science \& Technology}, vol.~13, no.~11, p.~28, Nov. 2024. [Online]. Available: \url{https://doi.org/10.1167/tvst.13.11.28}
\BIBentrySTDinterwordspacing

\bibitem{fauvel_human---loop_2022}
\BIBentryALTinterwordspacing
T.~Fauvel and M.~Chalk, ``\BIBforeignlanguage{en}{Human-in-the-loop optimization of visual prosthetic stimulation},'' \emph{\BIBforeignlanguage{en}{Journal of Neural Engineering}}, vol.~19, no.~3, p. 036038, Jun. 2022, publisher: IOP Publishing. [Online]. Available: \url{https://dx.doi.org/10.1088/1741-2552/ac7615}
\BIBentrySTDinterwordspacing

\bibitem{Brindley1968}
\BIBentryALTinterwordspacing
G.~S. Brindley and W.~S. Lewin, ``The sensations produced by electrical stimulation of the visual cortex,'' \emph{The Journal of Physiology}, vol. 196, no.~2, pp. 479--493, 1968. [Online]. Available: \url{https://physoc.onlinelibrary.wiley.com/doi/abs/10.1113/jphysiol.1968.sp008519}
\BIBentrySTDinterwordspacing

\bibitem{Dobelle1974}
\BIBentryALTinterwordspacing
W.~H. Dobelle and M.~G. Mladejovsky, ``Phosphenes produced by electrical stimulation of human occipital cortex, and their application to the development of a prosthesis for the blind,'' \emph{The Journal of Physiology}, vol. 243, no.~2, pp. 553--576, 1974. [Online]. Available: \url{https://physoc.onlinelibrary.wiley.com/doi/abs/10.1113/jphysiol.1974.sp010766}
\BIBentrySTDinterwordspacing

\bibitem{nanduri_frequency_2012}
D.~Nanduri, I.~Fine, A.~Horsager, G.~M. Boynton, M.~S. Humayun, R.~J. Greenberg, and J.~D. Weiland, ``Frequency and amplitude modulation have different effects on the percepts elicited by retinal stimulation,'' \emph{Invest Ophthalmol Vis Sci}, vol.~53, no.~1, pp. 205--14, Jan. 2012.

\bibitem{horsager_predicting_2009}
A.~Horsager, S.~H. Greenwald, J.~D. Weiland, M.~S. Humayun, R.~J. Greenberg, M.~J. McMahon, G.~M. Boynton, and I.~Fine, ``Predicting visual sensitivity in retinal prosthesis patients,'' \emph{Invest Ophthalmol Vis Sci}, vol.~50, no.~4, pp. 1483--91, Apr. 2009.

\bibitem{prashnani_pieapp_2018}
\BIBentryALTinterwordspacing
E.~Prashnani, H.~Cai, Y.~Mostofi, and P.~Sen, ``{PieAPP}: {Perceptual} {Image}-{Error} {Assessment} {Through} {Pairwise} {Preference},'' in \emph{2018 {IEEE}/{CVF} {Conference} on {Computer} {Vision} and {Pattern} {Recognition}}, Jun. 2018, pp. 1808--1817, conference Name: 2018 IEEE/CVF Conference on Computer Vision and Pattern Recognition (CVPR) ISBN: 9781538664209 Place: Salt Lake City, UT Publisher: IEEE. [Online]. Available: \url{https://ieeexplore.ieee.org/document/8578292/}
\BIBentrySTDinterwordspacing

\bibitem{turkstra_assistive_2025}
\BIBentryALTinterwordspacing
L.~M. Turkstra, T.~Bhatia, A.~Van~Os, and M.~Beyeler, ``\BIBforeignlanguage{en}{Assistive technology use in domestic activities by people who are blind},'' \emph{\BIBforeignlanguage{en}{Scientific Reports}}, vol.~15, no.~1, p. 7486, Mar. 2025, publisher: Nature Publishing Group. [Online]. Available: \url{https://www.nature.com/articles/s41598-025-91755-w}
\BIBentrySTDinterwordspacing

\bibitem{granley_beyond_2024}
\BIBentryALTinterwordspacing
J.~Granley, G.~Pogoncheff, A.~Rodil, L.~Soo, L.~M. Turkstra, L.~G. Nadolskis, A.~A. Saez, C.~S. Sanchez, E.~F. Jover, and M.~Beyeler, ``Beyond {Sight}: {Probing} {Alignment} {Between} {Image} {Models} and {Blind} {V1},'' Mar. 2024, arXiv:2403.12990 [q-bio]. [Online]. Available: \url{http://arxiv.org/abs/2403.12990}
\BIBentrySTDinterwordspacing

\bibitem{Okorokova}
E.~Okorokova, Q.~He, and S.~Bensmaia, ``Biomimetic encoding model for restoring touch in bionic hands through a nerve interface,'' \emph{Journal of Neural Engineering}, vol.~15, 09 2018.

\bibitem{Dorman}
M.~Dorman, A.~Spahr, P.~Loizou, C.~Dana, and J.~Schmidt, ``\BIBforeignlanguage{English (US)}{Acoustic simulations of combined electric and acoustic hearing (eas)},'' \emph{\BIBforeignlanguage{English (US)}{Ear and hearing}}, vol.~26, no.~4, pp. 371--380, Aug. 2005, copyright: Copyright 2008 Elsevier B.V., All rights reserved.

\bibitem{drakopoulos_dnn-based_2023}
\BIBentryALTinterwordspacing
F.~Drakopoulos, A.~Van Den~Broucke, and S.~Verhulst, ``A {DNN}-{Based} {Hearing}-{Aid} {Strategy} {For} {Real}-{Time} {Processing}: {One} {Size} {Fits} {All},'' in \emph{{ICASSP} 2023 - 2023 {IEEE} {International} {Conference} on {Acoustics}, {Speech} and {Signal} {Processing} ({ICASSP})}, Jun. 2023, pp. 1--5, iSSN: 2379-190X. [Online]. Available: \url{https://ieeexplore.ieee.org/abstract/document/10094887}
\BIBentrySTDinterwordspacing

\bibitem{lesica_harnessing_2021}
\BIBentryALTinterwordspacing
N.~A. Lesica, N.~Mehta, J.~G. Manjaly, L.~Deng, B.~S. Wilson, and F.-G. Zeng, ``\BIBforeignlanguage{en}{Harnessing the power of artificial intelligence to transform hearing healthcare and research},'' \emph{\BIBforeignlanguage{en}{Nature Machine Intelligence}}, vol.~3, no.~10, pp. 840--849, Oct. 2021, publisher: Nature Publishing Group. [Online]. Available: \url{https://www.nature.com/articles/s42256-021-00394-z}
\BIBentrySTDinterwordspacing

\bibitem{rao_brain_2023}
\BIBentryALTinterwordspacing
R.~P.~N. Rao, ``\BIBforeignlanguage{en}{Brain {Co}-processors: {Using} {AI} to {Restore} and {Augment} {Brain} {Function}},'' in \emph{\BIBforeignlanguage{en}{Handbook of {Neuroengineering}}}, N.~V. Thakor, Ed.\hskip 1em plus 0.5em minus 0.4em\relax Singapore: Springer Nature, 2023, pp. 1225--1260. [Online]. Available: \url{https://doi.org/10.1007/978-981-16-5540-1_32}
\BIBentrySTDinterwordspacing

\end{thebibliography}

\end{document}